\documentclass{amsart}

\usepackage{hyperref}

\usepackage{amssymb}
\usepackage{amsmath}
\usepackage{amsthm}
\usepackage{enumerate}
\usepackage{graphicx}
\usepackage{color}

\theoremstyle{plain}

\makeatletter
\newtheorem*{rep@theorem}{\rep@title}
\newcommand{\newreptheorem}[2]{%
\newenvironment{rep#1}[1]{%
 \def\rep@title{#2 \ref{##1}}%
 \begin{rep@theorem}}%
 {\end{rep@theorem}}}
\makeatother

\newtheorem{theorem}{Theorem}[section]
\newreptheorem{theorem}{Theorem}

\newtheorem{lemma}[theorem]{Lemma}

\newtheorem{conjecture}[theorem]{Conjecture}

\theoremstyle{definition}

\theoremstyle{remark}
\newtheorem*{remark}{Remark}

\begin{document}

\title{A combinatorial conjecture from PAC-Bayesian machine learning}

\date{\today}

\author[M. Younsi]{Malik Younsi}
\address{Department of Mathematics, University of Hawaii Manoa, Honolulu, HI 96822, USA.}
\email{malik.younsi@gmail.com}
\author[A. Lacasse]{Alexandre Lacasse}
\address{Coveo Solutions Inc., Qu\'ebec, Qu\'ebec G1W 2K7, Canada.}
\email{alacasse@coveo.com}

\thanks{The first author was supported by NSF Grant DMS-1758295.}

\keywords{Combinatorial identities, Binomial sums, multinomial sums, PAC-Bayesian machine learning.}
%\subjclass[2010]{primary 30-02 ; secondary 30E20.}

\begin{abstract}
We present a proof of a combinatorial conjecture from the second author's Ph.D. thesis \cite{LAC}. The proof relies on binomial and multinomial sums identities. We also discuss the relevance of the conjecture in the context of PAC-Bayesian machine learning.
\end{abstract}

\maketitle

\section{PAC-Bayes bounds in machine learning.}
\label{sec1}

In machine learning, ensemble methods are used to build stronger models by combining several base models. In the setting of classification problems, the model, called \textit{classifier}, can be combined using a weighted majority vote of the base classifiers. Examples of learning algorithms built using such classifiers include \textit{Bagging} \cite{BRE} and \textit{Boosting} \cite{FRS}.

The PAC-Bayes theorem, introduced by McAllester in the 1990's (\cite{MCA1}, \cite{MCA2}, \cite{MCA3}), provides bounds on the expectation of the risk of the base classifiers among a majority vote classifier. These bounds are computed from the empirical risk of the majority classifier made during the training phase of a learning algorithm, and can then be used to provide guarantees on existing algorithms or to design new ones.

The version of the PAC-Bayes theorem in \cite{GLL} involves a function $\xi: \mathbb{N} \to \mathbb{R}$ defined by
$$\xi(m):= \sum_{k=0}^m \binom{m}{k} \left( \frac{k}{m} \right)^k \left( 1-\frac{k}{m} \right)^{m-k}$$
and improves upon other versions, such as the ones in \cite{LAN} and \cite{MAU} for instance, which provide the bounds $m+1$ and $2\sqrt{m}$ respectively using approximations. The version in \cite{GLL}, on the other hand, is based on a direct calculation in order to obtain $\xi(m)$.

Bounds on the risk of the majority vote itself can be deduced from the bounds given by the PAC-Bayes theorem. Such bounds, however, turn out to be greater than the bounds on the base voters. In order to circumvent this issue and obtain better bounds, a new version of the PAC-Bayes theorem was given in \cite{LLM}. This new theorem can be seen as a two-dimensional version, since it gives bounds simultaneously on the expectation of the joint errors \textit{and} on the joint success of pairs of voters in a majority vote. As shown in \cite{LLM}, this approach does indeed lead to better bounds on the risk of the majority vote. The method of \cite{LLM} is based on techniques from \cite{GLL} and involves another combinatorial sum function $\xi_2:\mathbb{N} \to \mathbb{R}$ defined by
$$\xi_2(m):= \sum_{j=0}^m \sum_{k=0}^{m-j} \binom{m}{j} \binom{m-j}{k} \left( \frac{j}{m} \right)^j \left( \frac{k}{m} \right)^k \left( 1-\frac{j}{m}-\frac{k}{m} \right)^{m-j-k}.$$
The function $\xi_2$ was also used in \cite{GLLMR} and \cite{LAC}.

\section{A combinatorial conjecture.}

In \cite{LAC}, the second author posed the following conjecture, based on numerical evidence.

\begin{conjecture}[Lacasse \cite{LAC}]
\label{ConjectureLacasse}
For every $m \in \mathbb{N}$, we have
$$\xi_2(m)=\xi(m)+m.$$
\end{conjecture}

In this paper, we present a proof of Conjecture \ref{ConjectureLacasse} due to the first author that appeared in a an unpublished manuscript on the arXiv \cite{YOU}. The proof uses binomial and multinomial sums identities in order to obtain simpler expressions for the functions $\xi$ and $\xi_2$. It has been cited in several publications related to machine learning, such as \cite{FD}, \cite{GER}, \cite{GLLMR}, \cite{ONE}, \cite{ONE2}, \cite{ONE3}, \cite{OAR}, \cite{SS}. We also mention that although other proofs of Conjecture \ref{ConjectureLacasse} were subsequently obtained (see e.g. \cite{CPY}, \cite{GES}, \cite{PRO}, \cite{SUN}), the proof we present remains, as far as we know, the only one providing equivalent expressions for the functions $\xi$ and $\xi_2$ that are simpler and more convenient from a numerical perspective. Furthermore, none of the aforementioned papers \cite{CPY}, \cite{GES}, \cite{PRO}, \cite{SUN} discuss the context and relevance of Conjecture \ref{ConjectureLacasse} in the theory of machine learning.

\section{Proof of Conjecture \ref{ConjectureLacasse}}

In this section, we present the proof of Conjecture \ref{ConjectureLacasse} from \cite{YOU}.

\subsection{An equivalent formulation.}

It will be more convenient to express the conjecture in terms of binomial and multinomial type sums. We therefore introduce the following functions:
$$\gamma(m):= m^m \xi(m) = \sum_{k=0}^m \binom{m}{k} k^k (m-k)^{m-k} \qquad (m \in \mathbb{N})$$
and
$$\gamma_2(m):= m^m \xi_2(m) = \sum_{j=0}^m \sum_{k=0}^{m-j} \binom{m}{j} \binom{m-j}{k} j^j k^k (m-j-k)^{m-j-k} \qquad (m \in \mathbb{N}).$$

Note that Conjecture \ref{ConjectureLacasse} is equivalent to the identity
$$\gamma_2(m)-\gamma(m) = m^{m+1} \qquad (m \in \mathbb{N}).$$

It turns out that the functions $\gamma$ and $\gamma_2$ have considerably simpler expressions.

\begin{lemma}
\label{lem1}

For $m \in \mathbb{N}$, we have

$$\gamma(m) = \sum_{j=0}^m m^j \frac{m!}{j!}$$
and
$$\gamma_2(m) = \sum_{j=0}^m m^{m-j} \binom{m}{j} (j+1)!.$$

\end{lemma}

Assuming Lemma \ref{lem1}, we can now prove Conjecture \ref{ConjectureLacasse}.

\begin{proof}
The proof consists of an elementary calculation. We have

\begin{eqnarray*}
\gamma_2(m) - \gamma(m) &=& \sum_{j=0}^m m^{m-j} \binom{m}{j} (j+1)! - \sum_{j=0}^m m^j \frac{m!}{j!}\\
&=& \sum_{k=0}^m m^k  \binom{m}{m-k} (m-k+1)! - \sum_{j=0}^m m^j \frac{m!}{j!}\\
&=& \sum_{k=0}^m m^k  \frac{m!}{k!}  (m-k+1) - \sum_{j=0}^m m^j \frac{m!}{j!}\\
&=& \sum_{k=0}^m m^k \frac{m!}{k!} (m-k)\\
&=& m\sum_{k=0}^m m^k \frac{m!}{k!} - \sum_{k=0}^m km^k \frac{m!}{k!}\\
&=& \sum_{j=1}^{m+1} m^j \frac{m!}{(j-1)!} - \sum_{k=1}^m m^k \frac{m!}{(k-1)!}\\
&=& m^{m+1}
\end{eqnarray*}
as required.
\end{proof}

\subsection{Proof of Lemma \ref{lem1}.}

For the proof of Lemma \ref{lem1}, we need a binomial sum identity first proved by Abel as well as its generalization to the multinomial case by Hurwitz.

More precisely, for $m \in \mathbb{N}, x,y \in \mathbb{R}, p,q \in \mathbb{Z}$, define
$$A_m(x,y;p,q):= \sum_{k=0}^m \binom{m}{k} (x+k)^{k+p}(y+m-k)^{m-k+q}.$$

Note that the special case $p=0$, $q=-1$, $y \neq 0$ corresponds to the classical Abel binomial theorem:

$$A_m(x,y;0,-1) = \frac{1}{y} (x+y+m)^m.$$

Moreover, we have
\begin{equation}
\label{eq1}
A_m(0,0;0,0) = \gamma(m) \qquad (m \in \mathbb{N}).
\end{equation}

The other function $\gamma_2$ can be expressed in terms of a multinomial version of $A_m$. More precisely, for $x_1, \dots, x_n \in \mathbb{R}$ and $p_1, \dots, p_n \in \mathbb{Z}$, define
$$B_m(x_1, \dots, x_n; p_1, \dots, p_n):= \sum \frac{m!}{k_1! \cdots k_n!} \prod_{j=1}^n (x_j+k_j)^{k_j+p_j},$$
where the sum is taken over all non-negative integers $k_1, \dots, k_n$ such that $k_1+ \dots + k_n=m.$

A simple calculation then shows that

\begin{equation}
\label{eq2}
B_m(0,0,0;0,0,0) = \gamma_2(m) \qquad (m \in \mathbb{N}).
\end{equation}

We can now proceed with the proof of Lemma \ref{lem1}.

\begin{proof}
In \cite[p.21]{RIO}, we find the following identity:

$$
A_m(x,y;0,0) = \sum_{k=0}^m \binom{m}{k} k! \, (x+y+m)^{m-k}
$$

Setting $x=y=0$ gives

$$
\gamma(m)=A_m(0,0;0,0) = \sum_{k=0}^m \binom{m}{k} k! \, m^{m-k}= \sum_{j=0}^m \frac{m!}{j!}m^j,
$$
where we used Equation (\ref{eq1}). This proves the first equality.

 For the second equality, we use \cite[Equation (35), p.25]{RIO}:

$$
B_m(x_1, \dots, x_n;0, \dots, 0) = \sum_{k=0}^m \binom{m}{k} (x_1+ \dots x_n + m)^{m-k} \alpha_k(n-1)
$$
where
$$\alpha_k(r):= \frac{(r+k-1)!}{(r-1)!}.$$
Note that if $n=3$, then $\alpha_k(n-1) = (k+1)!$. Setting $x_1=x_2=x_3=0$ then gives
$$
\gamma_2(m) = B_m(0,0,0;0,0,0) = \sum_{k=0}^m \binom{m}{k} m^{m-k} (k+1)!,
$$
where we used Equation (\ref{eq2}). This completes the proof of the lemma.
\end{proof}

\subsection{Conclusion.}

We now summarize what we have proved in the following theorem.

\begin{theorem}
\label{mainthm}
For $m \in \mathbb{N}$, define
$$\xi(m):= \sum_{k=0}^m \binom{m}{k} \left( \frac{k}{m} \right)^k \left( 1-\frac{k}{m} \right)^{m-k}$$
and
$$\xi_2(m):= \sum_{j=0}^m \sum_{k=0}^{m-j} \binom{m}{j} \binom{m-j}{k} \left( \frac{j}{m} \right)^j \left( \frac{k}{m} \right)^k \left( 1-\frac{j}{m}-\frac{k}{m} \right)^{m-j-k}.$$

Then we have
$$\xi(m) = \frac{1}{m^m} \sum_{j=0}^m m^j \frac{m!}{j!} \qquad (m \in \mathbb{N})$$
and
$$\xi_2(m) = \frac{1}{m^m}\sum_{j=0}^m m^{m-j} \binom{m}{j} (j+1)! \qquad (m \in \mathbb{N}).$$

Furthermore,
$$\xi_2(m) = \xi(m)+m \qquad (m \in \mathbb{N}).$$
\end{theorem}

\begin{remark}
Theorem \ref{mainthm} not only proves Conjecture \ref{ConjectureLacasse}, but also gives simpler expressions for the functions $\xi$ and $\xi_2$ that are more convenient from a numerical perspective. As discussed in Section \ref{sec1}, this might be of interest for the computation of PAC-Bayes bounds in machine learning.
\end{remark}

\bibliographystyle{amsplain}

\begin{thebibliography}{99}

\bibitem{BRE}
L. Breiman,
Bagging predictors,
\textsl{Mach. Learn.},
\textbf{24} (1996),
123--140.

\bibitem{CPY}
W.Y.C. Chen, J.F.F. Peng, H.R.L. Yang,
Decomposition of triply rooted trees,
\textsl{Electron. J. Combin.},
\textbf{20} (2013),
10pp.

\bibitem{FD}
E.B. Fox, D.B. Dunson,
Bayesian Nonparametric Covariance Regression,
\textsl{Journal of Machine Learning Research},
\textbf{16} (2015),
2501--2542.

\bibitem{FRS}
Y. Freund and R.E. Schapire,
A decision-theoretic generalization of on-line learning and application to boosting,
In \textsl{Proceedings of the Second European Conference on Computational Learning Theory},
EuroCOLT ’95,
1995,
23--37.

\bibitem{GES}
I.M. Gessel,
Lagrange inversion,
\textsl{J. Combin. Theory Ser. A},
\textbf{144} (2016),
212--249.

\bibitem{GER}
P. Germain,
G\'en\'eralisations de la th\'eorie PAC-bay\'esienne pour l'apprentissage inductif, l'apprentissage transductif et l'adaptation de domaine,
Ph.D. Thesis, Universit\'e Laval,
2015.

\bibitem{GLL}
P. Germain, A. Lacasse, F. Laviolette, M. Marchand,
PAC-Bayesian Learning of Linear Classifiers,
In \textsl{Proceedings of the 26th Annual International Conference on Machine Learning},
ICML ’09,
2009,
353--360.


\bibitem{GLLMR}
P. Germain, A. Lacasse, F. Laviolette, M. Marchand, J.-F. Roy,
Risk Bounds for the Majority Vote: From a PAC-Bayesian Analysis to a Learning Algorithm,
\textsl{Journal of Machine Learning Research},
\textbf{16} (2015),
787--860.



\bibitem{LAC}
A. Lacasse,
\textsl{Bornes PAC-Bayes et algorithmes d'apprentissage},
Ph.D. Thesis, Universit\'e Laval,
2010.

\bibitem{LLM}
A. Lacasse, F. Laviolette, M. Marchand, P. Germain, N. Usunier,
PAC-Bayes Bounds for the Risk of the Majority Vote and the Variance of the Gibbs Classifier,
In \textsl{Advances in Neural Information Processing Systems 19},
2009,
769--776.

\bibitem{LAN}
J. Langford,
Tutorial on Practical Prediction Theory for Classification,
\textsl{J. Mach. Learn. Res.},
\textbf{6} (2005),
273--306.

\bibitem{MAU}
A. Maurer,
A note on the PAC bayesian theorem,
\textsl{CoRR},
2004.

\bibitem{MCA1}
D.A. McAllester,
Some pac-bayesian theorems,
In \textsl{Proceedings of the Eleventh Annual Conference on Computational Learning Theory},
COLT’ 98,
1998,
230--234.

\bibitem{MCA2}
D.A. McAllester,
Some PAC-Bayesian theorems,
\textsl{Machine Learning},
\textbf{37} (1999),
355--363.

\bibitem{MCA3}
D.A. McAllester,
Pac-bayesian stochastic model selection,
\textsl{Mach. Learn.},
\textbf{51} (2003),
5--21.

\bibitem{ONE}
L. Oneto,
Model selection and error estimation without the agonizing pain,
\textsl{WIREs Data Mining and Knowledge Discovery},
\textbf{8} (2018).

\bibitem{ONE2}
L. Oneto,
PAC-Bayes Theory,
\textsl{Model Selection and Error Estimation in a Nutshell},
(2019),
75--86.

\bibitem{ONE3}
L. Oneto,
The Five W of MS and EE,
\textsl{Model Selection and Error Estimation in a Nutshell},
(2019),
5--11.



\bibitem{OAR}
L. Oneto, D. Anguita, S. Ridella,
PAC-bayesian analysis of distribution dependent priors: Tighter risk bounds and stability analysis,
\textsl{Pattern Recognition Letters},
\textbf{80} (2016),
200--207.

\bibitem{PRO}
H. Prodinger,
An identity conjectured by Lacasse via the tree function,
\textsl{Electron. J. Combin.},
\textbf{20} (2013),
3pp.

\bibitem{RIO}
J. Riordan,
\textsl{Combinatorial Identities},
Robert E. Krieger Publishing Co., New York,
1968.

\bibitem{SS}
D.M. Smith, G. Smith,
Tight Bounds on Information Leakage from Repeated Independent Runs,
\textsl{2017 IEEE 30th Computer Security Foundations Symposium (CSF)}.


\bibitem{SUN}
Y. Sun,
A simple proof of an identity of Lacasse,
\textsl{Electron. J. Combin.},
\textbf{20} (2013),
3pp.




\bibitem{YOU}
M. Younsi,
Proof of a combinatorial conjecture coming from the PAC-Bayesian machine learning theory,
arXiv:1209.0824




\end{thebibliography}

\end{document}